\documentclass[conference]{IEEEtran}

\newcommand{\subparagraph}{}

\usepackage{amsmath}
\usepackage{amsfonts}
\usepackage{graphics, graphicx}
\usepackage{cite}
\usepackage{multirow}
\usepackage{url}
\usepackage{enumerate}
\usepackage{etoolbox}
\usepackage{numprint}
\usepackage{titlesec}
\usepackage{bm}
\usepackage{algorithm, algorithmic}

\npthousandsep{,}\npthousandthpartsep{}\npdecimalsign{.}

\makeatletter
\newcommand{\removelatexerror}{\let\@latex@error\@gobble}
\makeatother

\DeclareMathOperator*{\argmax}{argmax}

\title{Detecting Patch Adversarial Attacks with Image Residuals}

\author{\IEEEauthorblockN{Marius Arvinte, Ahmed H. Tewfik, and Sriram Vishwanath}
	\IEEEauthorblockA{Department of Electrical and Computer Engineering\\
		University of Texas at Austin\\
		Austin, Texas 78712\\
		Email: arvinte@utexas.edu}}

\makeatletter
\patchcmd{\@maketitle}
{\addvspace{0.8\baselineskip}\egroup}
{\addvspace{-0.4\baselineskip}\egroup}
{}
{}
\makeatother

\begin{document}
	
	\maketitle
	
	\begin{abstract}
		We introduce an adversarial sample detection algorithm based on image residuals, specifically designed to guard against patch-based attacks. The image residual is obtained as the difference between an input image and a denoised version of it, and a discriminator is trained to distinguish between clean and adversarial samples. More precisely, we use a wavelet domain algorithm for denoising images and demonstrate that the obtained residuals act as a \textit{digital fingerprint} for adversarial attacks. To emulate the limitations of a physical adversary, we evaluate the performance of our approach against localized (patch-based) adversarial attacks, including in settings where the adversary has complete knowledge about the detection scheme. Our results show that our proposed method generalizes to stronger attacks and reduces the success rate (conversely, increases the computational effort) of an adaptive attacker.
	\end{abstract}
	
	\section{Introduction}
	\label{section_intro}
	In the past decade, deep neural networks (DNNs) have been demonstrated to match and surpass human performance on image classification tasks and have become ubiquitous in machine learning. At the same time, DNNs have been shown to be very fragile to \textit{adversarial examples} \cite{goodfellow2014explaining}, in which a malicious user perturbs natural images such that they are misclassified by the model. A growing body of research investigates adversarial defense methods and their shortcomings \cite{athalye2018obfuscated, carlini2017adversarial}. Anomaly detection is one of two currently widely investigated lines of defense and is differentiated (but not necessarily in opposition) from robustness, where the goal is to recover the ground truth from the perturbed test sample. In our work, we develop a solution for the detection problem, with applications in security systems that can alert human operators and prompt intervention (e.g., security cameras, autonomous driving) when  abnormal samples are detected.
	
	Recent work has investigated adversarial patch attacks as a step towards physically realizable and robust threat models \cite{46561, karmon2018lavan}. Through careful digital design, physical adversarial patches represent a currently unsolved security threat. We focus on the task of \textit{detecting} patch attacks at test time through the use of a detection block trained on a small subset of samples within the adversary's model. We introduce multiple threat models that follow the taxonomy in \cite{carlini2017adversarial} with the common characteristic that they are all confined to a digital patch attack: given a clean test sample, the adversary can only modify a contiguous, rectangular region of it, with a size up to $6.25\%$ of the image. We consider both the cases of norm-bounded and unbounded adversaries operating on the patch.
	
	Central to our approach is the idea of detecting adversarial samples based on image residuals, obtained as the difference between an input image and a denoised version of it. These residuals are used to train a secondary, much smaller and heavily regularized detection neural network. We experimentally demonstrate that the proposed method is robust and generalized to different patch-based attacks, including much stronger than the ones used to train the detector network. We show that this generalization does not happen for state-of-the-art detection methods that are not specifically designed for patch threat models. Very recent work on defenses against adversarial patches focuses only on the robustness problem, not detection \cite{Chiang*2020Certified, alex2020derandomized} to produce certified guarantees. However, these approaches rely on a brute-force approach that requires additional complexity at inference time. Thus, there is a need of specialized solutions for detecting patch adversarial attacks, and we hope our work provides a baseline in this direction. Source code is available at \texttt{\url{https://github.com/mariusarvinte/wavelet-patch-detection}}.
	
	We carry out experiments on the CIFAR-10 dataset \cite{krizhevsky2009learning} with a VGG-19 \cite{simonyan2014very} deep convolutional classifier architecture and show that current state-of-the-art detection methods do not generalize to different patch attacks, while our proposed solution does. The performance of our scheme is evaluated against full- and limited-knowledge adversaries that attempt to bypass the issue of zero gradients coming from the non-differentiable nature of the wavelet denoising operator, as well as adversaries that perform a brute-force search for the best patch location.
	
	Summarized, our contributions are:
	\begin{itemize}
		\item We introduce an adversarial sample detection algorithm based on image residuals, obtained as the difference between an image and a wavelet-filtered version of it.
		\item We investigate the effectiveness of several patch adversarial attacks against our proposed detection scheme and two other existing detection schemes, showing that our approach generalizes where prior work does not.
		\item We show that our approach resists high-confidence transfer attacks, even when the attacker trains a substitute detector and lowers the success rate of an adaptive attack.
	\end{itemize}
	
	\section{Methodology}
	\subsection{Image Residuals}
	
	Let $F_i(x)$ denote the output probabilities of a deep neural network classifier $F(\cdot; \theta_F)$ with weights $\theta_F$, where $C$ is the number of classes, $i = 1, \dots, C$ and $x$ is an input image to the network, with its true label $y$. The network is trained to minimize its loss function $L(x, y; \theta_F)$; typically, this is chosen as the categorical cross-entropy between the predicted and true labels. Let $Z_i(x)$ denote the output logits of the network, where we assume that the last layer uses a softmax activation and the relationship $F_i(x) = \frac{\exp(Z_i(x))}{\sum_k \exp(Z_k(x))}$ holds. 
	
	The image residual of $x$ is defined as
	\begin{equation}
	R(x) = x - g(x; \sigma),
	\end{equation}
	
	\noindent where $g(x; \sigma)$ is a denoising operator that takes as input the image $x$ and produces a denoised version of it with the same dimensions. $g$ is adjustable by the parameter $\sigma$, which represents an estimate of the noise power in the image: the larger $\sigma$ is chosen, the more aggressive the denoising is, resulting in a smoother image.
	
	For the rest of our work, we choose $g$ to be a wavelet-based denoising operator. In particular, we use the adaptive Bayesian shrinkage algorithm \cite{chipman1997adaptive}, in which $\sigma$ plays the role of noise power. Once a residual is obtained, it is passed to the detector function $H(\cdot; \theta_H)$, parameterized by weights $\theta_H$. If an adversarial sample is detected, an alarm is triggered, otherwise, the most likely class is predicted.

	\subsection{Counteracting Blind Spots}
	Since our approach involves a block that relies on identifying anomalous high-frequency structure in images, we anticipate the following weakness: an attacker may employ very small patches intentionally, leaving no residual signature. An extreme case of this is the single-pixel attack \cite{su2019one}, which falls in our threat model as a patch of size $1 \times 1$. Consequently, we augment our detector by taking inspiration from the logit margin loss formulation in \cite{carlini2017adversarial} and the baseline out-of-distribution detection method introduced in \cite{hendrycks2016baseline}. Let $D_i(x), i \in \{ 1, \dots, C+1 \}$ be the logits of the joint classifier-detector, given by
	\begin{equation}
	D_i(x) = \left\{
	\begin{array}{ll}
	Z_i(x) &, \forall i \in 1, \dots, C \\
	(1 + Y(x)) \max_j Z_j(x) &, i = C + 1\\
	\end{array} 
	\right.
	.
	\label{eq:joint_logits}
	\end{equation}
	
	The two stage detection procedure is described in Table \ref{alg:two_stage_det}. The core idea is to augment the residual-based detection by requiring negatively labeled samples pass a confidence threshold in their predictions. The parameter $\kappa_{det}$ plays this role: a sample is declared non-adversarial only if it bypasses the detector and if the difference between the two highest prediction logits is at least $\kappa_{det}$, implying that we require natural images to be confidently classified.
	
	\begin{table}
		\caption{Two-Stage Detection Process}
		\label{alg:two_stage_det}
		\begin{algorithmic}[1]
			\STATE {\bfseries Input:} Logits $Z_i(x), Y(x)$
			\STATE {\bfseries Output:} Detection decision $\hat{y}_{det}$
			\STATE Compute logits $D_k(x), k=1, \dots, C+1$
			\STATE Compute $k^* = \argmax_k Z_k(x)$
			\IF{$\argmax_k D_k(x) == C+1$}
			\STATE $\hat{y}_{det} = 1$
			\ELSIF{$Z_{k^*} - \max_{k, k\ne k*} Z_k < \kappa_{det}$}
			\STATE $\hat{y}_{det} = 1$
			\ELSE
			\STATE $\hat{y}_{det} = 0$
			\ENDIF
		\end{algorithmic}
	\end{table}
	
	\subsection{Motivation}
		
	Recent work has investigated the frequency domain signature of adversarial attacks \cite{yin2019fourier} and has concluded that adversarial training \cite{madry2017towards}, one of the few unbroken defenses, decreases model sensitivity to high-frequency components of the input signal, implying that an undefended network will be perturbed by these components. We posit that this phenomenon is more pronounced for patch adversarial attacks, since distortions caused by the inserted patch will be visible in the residual $R$, at least for an adversary that is unaware of the existence of a detection scheme. At the same time, adaptive adversaries can try to evade detection by generating smooth patches, but this will reduce the effective dimension of the patch, ultimately lowering their success rate.
	
	\begin{figure}
		\begin{center}
			\includegraphics[width=0.8\columnwidth]{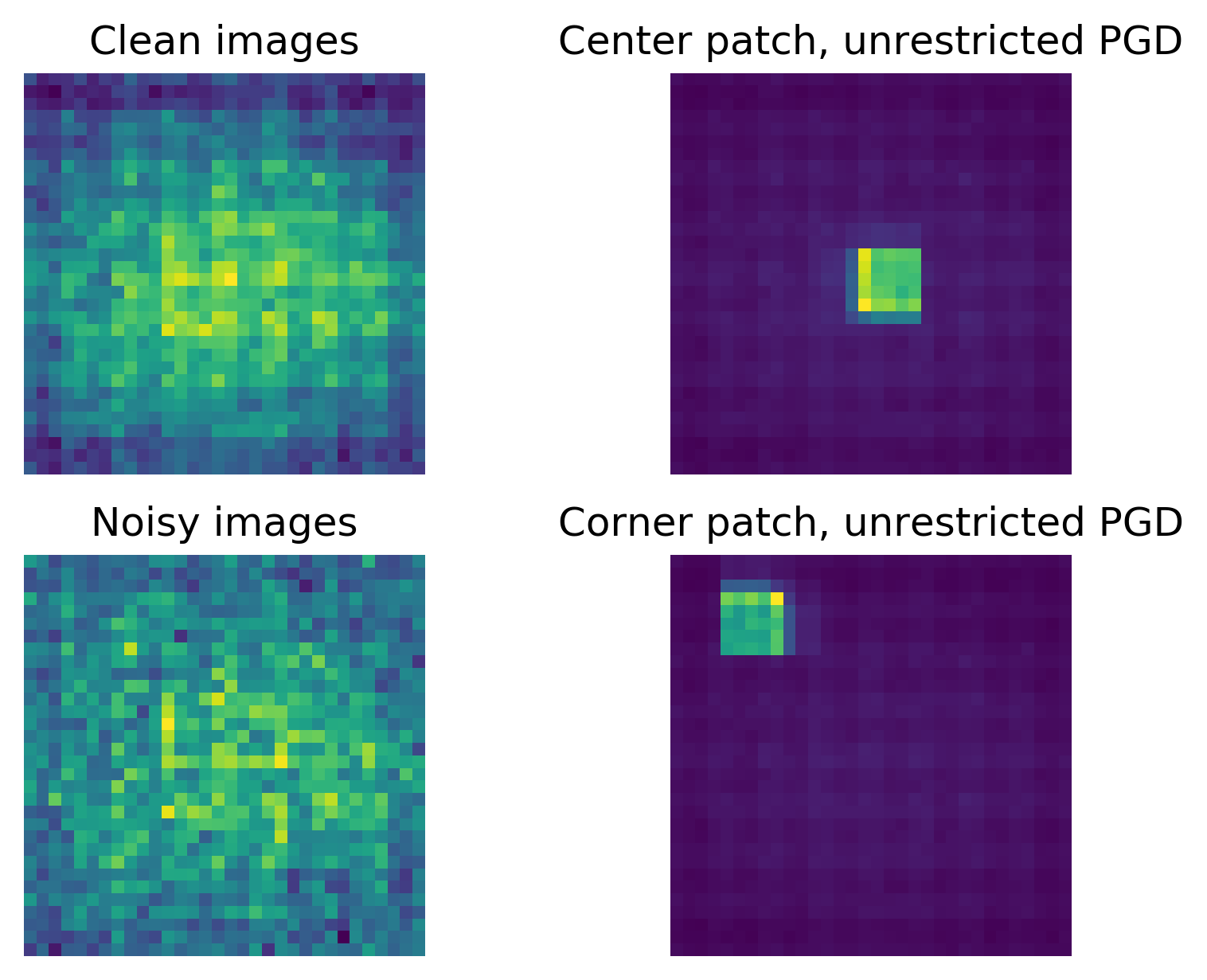}
			
			\caption{Average magnitude of residual $R$ for $200$ CIFAR-10 test images different conditions: clean, noisy or attacked with unrestricted patch PGD with step size $5$ and $100$ iterations, with restarts until the attack is successful. Magnitude averaged across all color channels.}
			
			\label{fig:avg_residual}
			
		\end{center}
	\end{figure}
	
	Figure \ref{fig:avg_residual} plots the magnitude of the residual averaged across $200$ test samples of the CIFAR-10 dataset in three cases: the original images, noisy versions of them, and unbounded projected gradient descent (PGD) adversarial samples corresponding to each original image. The average residual for the clean and noisy images presents high magnitude values around the center of the image, where edges are more likely to occur. For the adversarial examples, the patch is strongly localized in the residual, even when it is placed in the center, overtaking the residuals of semantic content in the image.
	
	\section{Adversarial Attacks}
	
	\subsection{Patch Adversarial Attacks}
	We use the same definition of patch adversarial attacks as \cite{karmon2018lavan}: the patch is characterized by a two-dimensional binary mask $m$, in which only a contiguous, rectangular, two-dimensional region $S$ satisfies $m(i) = 1, \forall i \in S$, otherwise $m(i) = 0$. A patch adversarial sample $x_{adv}$ is given by $x_{adv} = x + \delta$, where $\delta$ is the adversarial perturbation satisfying $\delta(S^c) = 0$. For most attacks, we assume that the attacker randomly samples a mask location in the image, but we also investigate the effectiveness of an adversary that searches across all possible mask locations.
	
	\subsubsection{Projected Gradient Descent}
	PGD \cite{madry2017towards} is an iterative attack that takes a series of steps in the gradient direction, each with size $\epsilon_{step}$, and projects the perturbations on the $\epsilon$-ball if their norm exceeds $\epsilon$. Additionally, PGD starts from a perturbed point around $x$. We use the untargeted, $L_{\infty}$ version of masked PGD, with the inner step given by
	\begin{equation}
	x^{t+1} = \Pi_{1}(x^t + \Pi_{\epsilon}(\epsilon_{step} \cdot m \odot \mathrm{sign}(\nabla_x L(x, y; \theta_F)))),
	\end{equation}
	
	\noindent where $\Pi_{\epsilon}$ projects each element to the interval $[0, \epsilon]$. Note that by choosing $\epsilon = 1$, this attack becomes unrestricted in the pixel space, assuming normalization of images to $[0, 1]$.
	
	\subsubsection{C\&W Attack \cite{carlini2017towards}}
	This is an optimization-based attack that reparameterizes the perturbation as $\delta = \frac{1}{2} (\tanh(m \odot w) + 1) - x$ and solves the optimization problem
	\begin{equation}
	\mathrm{min}_w \ \alpha || \delta ||_p + \lambda f(x + \delta),
	\label{eq:CW}
	\end{equation}
	
	\noindent where $f(x') = \max\{\max_{i \ne t}  Z_i(x') - Z_t(x'), -\kappa_{adv} \}$, $t$ is the target label different from the correct class and $\kappa_{adv}$ is a confidence parameter. The formulation in (\ref{eq:CW}) includes two hyper-parameters: $\lambda$ controls the trade-off between the distance penalty and misclassification, while $\alpha \in \{0, 1\}$ allows us to run an unrestricted attack. When $\alpha = 0$, the value of $\lambda$ does not matter in the optimization, except for influencing the learning rate, thus we set it to one.
	
	\subsubsection{Single-Pixel Attack \cite{su2019one}}
	 This is a powerful attack that only requires probability outputs of the model and does not use gradient information. The attack uses a differential evolution algorithm to perturb a single pixel that misclassifies the image.	We evaluate our performance on this attack since we expect it to be a blind spot for our proposed definition of image residuals, according to \cite{athalye2018obfuscated}.

	\begin{table*}[t]
		\caption{Black-box PGD detection performance in terms of area under the receiver operating curve (ROC-AUC), average precision (AP) and false positive rate at $95\%$ true positive rate for the proposed method, LID \cite{ma2018characterizing} and baseline \cite{hendrycks2016baseline}. Values are percentages.}
		\label{table:bbox}
		\centering
		\begin{tabular}{|l|c|c|c|c|r|}
			\hline
			\multirow{2}{*}{Attack} & \multirow{2}{*}{Attack Success Rate [\%]} & ROC-AUC $\uparrow$ & AP $\uparrow$ & FPR at TPR=$95 \%$ $\downarrow$ \\
			& & (Baseline/LID/Ours) & (Baseline/LID/Ours) & (Baseline/LID/Ours) \\
			\hline
			PGD $L_\infty, \epsilon=8/255$    & $15.89$ & $88.93$/$81.07$/$\bm{93.38}$ & $46.67$/$49.18$/$\bm{75.27}$ & $29.72$/$53.72$/$\bm{25.75}$ \\
			PGD $L_\infty, \epsilon=16/255$   & $25.55$ & $82.78$/$77.37$/$\bm{98.20}$ & $38.29$/$39.54$/$\bm{96.44}$ & $43.60$/$65.57$/$\bm{6.67}$ \\
			PGD $L_\infty, \epsilon=64/255$   & $62.77$ & $60.58$/$63.47$/$\bm{99.95}$ & $29.38$/$25.72$/$\bm{99.96}$ & $81.52$/$87.04$/$\bm{0.03}$ \\
			PGD, Unrestricted        		  & $82.21$ & $46.38$/$63.68$/$\bm{99.99}$ & $32.81$/$20.81$/$\bm{99.99}$ & $93.40$/$88.67$/$\bm{0.00}$ \\
			\hline
		\end{tabular}
		\vskip -0.1in
	\end{table*}
	
	\subsection{Existing Detection Algorithms}
	We compare the performance of our method with two existing algorithms. The baseline approach in \cite{hendrycks2016baseline} uses the probability of the predicted class as a discriminant for in- and out-of-distribution samples. While simple and not intended for adversarial sample detection, we borrow from this idea to impose that clean samples pass a confidence limit, as previously described. This method does not require training.
		
	Local Intrinsic Dimensionality (LID) \cite{ma2018characterizing} is a powerful detection method against non-adaptive adversaries that extracts a set of statistics for each test sample by computing distances to $k$-nearest neighbors in the training set. This method requires training on adversarial samples and claims generalization properties, thus we choose it as a comparison.

	Our primary purpose of comparing to these methods is to show that when faced against a patch black-box adversary, their detection performance degrades either against a novel type of attack or larger perturbations. Note that, to the best of our knowledge, there is only one other prior work that explicitly targets adversarial patch detection \cite{chou2018sentinet}, but their source code is not publicly available, thus we omit it from comparison. 
	
	\section{Performance Results}
	
	\subsection{Training Details}
	We train a VGG-19 classifier on the CIFAR-10  dataset, retain $10\%$ of the training samples for validation, and obtain a clean test accuracy of $89.95\%$. We further split our original validation set in a new training and validation set and use the training set to train $H$ and the validation set to pick the best wavelet denoising parameter $\sigma$. We perform a hyper-parameter search for $\sigma$ and find that $0.05$ offers the best validation performance. The training data for our detector consists of adversarial samples generated with an untargeted PGD attack with strength $\epsilon \in \{8, 16\} / 255$, retaining only the successful attacks. The negative class also includes noisy versions of the training samples, with discrete uniform noise between $[-3, 3]$ (before any scaling) added independently on each pixel value. The negative class validation data for the detector consists of clean and noisy images. The positive class validation data consists of successful adversarial samples generated with an untargeted patch PGD attack with strength $\epsilon = 64/255$ run for $100$ steps.
	
	The patch location is randomly selected as a rectangle with sides between $[4, 8]$ pixels and placed randomly at uniform in a location in the sample, such that the entire patch is present in the image. The architecture of the detector $H$ is a three hidden layer neural network, with two convolutional layers and one fully-connected layer. We use $L_2$ weight regularization of $5 \times 10^{-3}$ during training to avoid over-fitting the detector to the training data. We use values for $\kappa_{det} \in \{0, 3\}$ throughout our experiments.

	\subsection{Black-Box Attacks}
	All attacks in this section and future sections, except for the single-pixel attack, have complete knowledge about the classifier architecture and weights $\theta_F$, but are unaware that there is a detection method in place.
	
	\subsubsection{PGD Attack}
	
	We generate $128$ patch locations at random and for each location we attack $256$ randomly chosen test images. We compare the performance of our algorithm with LID and the baseline approach against a black-box PGD adversary. We perform a parameter search to find the best $k_{LID}$ (number of nearest neighbors) and $\sigma_{LID}$ values, using a batch size of $200$ samples and the same training and validation data used for our approach. The average performance results are shown in Table \ref{table:bbox}, where it can be seen that our proposed approach has better generalization properties when testing on different attack types and strengths. In particular, previous methods fail to identify the localized changes introduced by an adversary and exhibit a very high false positive rate. Our method with $\kappa_{det} = 0$ shows an opposite trend against the others: weaker attacks are harder to detect. For fair comparison, we also include the success rate of the PGD attack, where it can be seen that it is much lower for a norm-bounded restricted adversary -- thus the absolute number of missed detections is also lower.
	
	\subsubsection{C\&W Attack}
	
	For the norm-restricted C\&W attack, we perform a binary search for $\lambda$ in the range $[10^{-2}, 10^{10}]$. In both attacks, a square patch of size $6\times 6$ is randomly placed at a location of the image, and we optimize the objective in (\ref{eq:CW}) with an Adam optimizer, running for $10000$ iterations each step of the binary search.	We pick $256$ correctly classified images from the test set, and run a targeted attack towards a random class different that the ground truth. For all images, we test $40$ patch locations. We use $\kappa_{det} = 3$ as a confidence threshold. The results are summarized in Table \ref{table:bbox_cw}. We note that constraining the patch attack implicitly helps it bypass detection, since a more blended patch is generated, and our method explicitly relies on the saliency of the perturbed region. For attacks that bypass detection, the average $L_2$ norm of the perturbation is $0.247$ and $0.3$ for the restricted and unrestricted attacks, respectively.
	
	\begin{table}[t]
		\caption{Success Rates (on classifier) and detection accuracy of the proposed method for a norm-restricted and unconstrained black-box C\&W adversary with $\kappa_{det} = 3$ and $\kappa_{adv} = 3$.}
		\label{table:bbox_cw}
		\vskip 0.15in
		\begin{center}

					\begin{tabular}{|l|c|c|c|r|}
						\hline
						Attack & C\&W $L_2$ & C\&W, Unrestricted  \\
						\hline
						Average Success Rate  & $50.1$\% & $46.3$\% \\
						Det. Accuracy & $85.62$\% & $98.55$\% \\

						\hline
					\end{tabular}
		\end{center}
		\vskip -0.1in
	\end{table}
	
	\subsubsection{Brute-Force PGD Attack}
	
\begin{figure}
		\begin{center}
			\includegraphics[width=0.8\columnwidth]{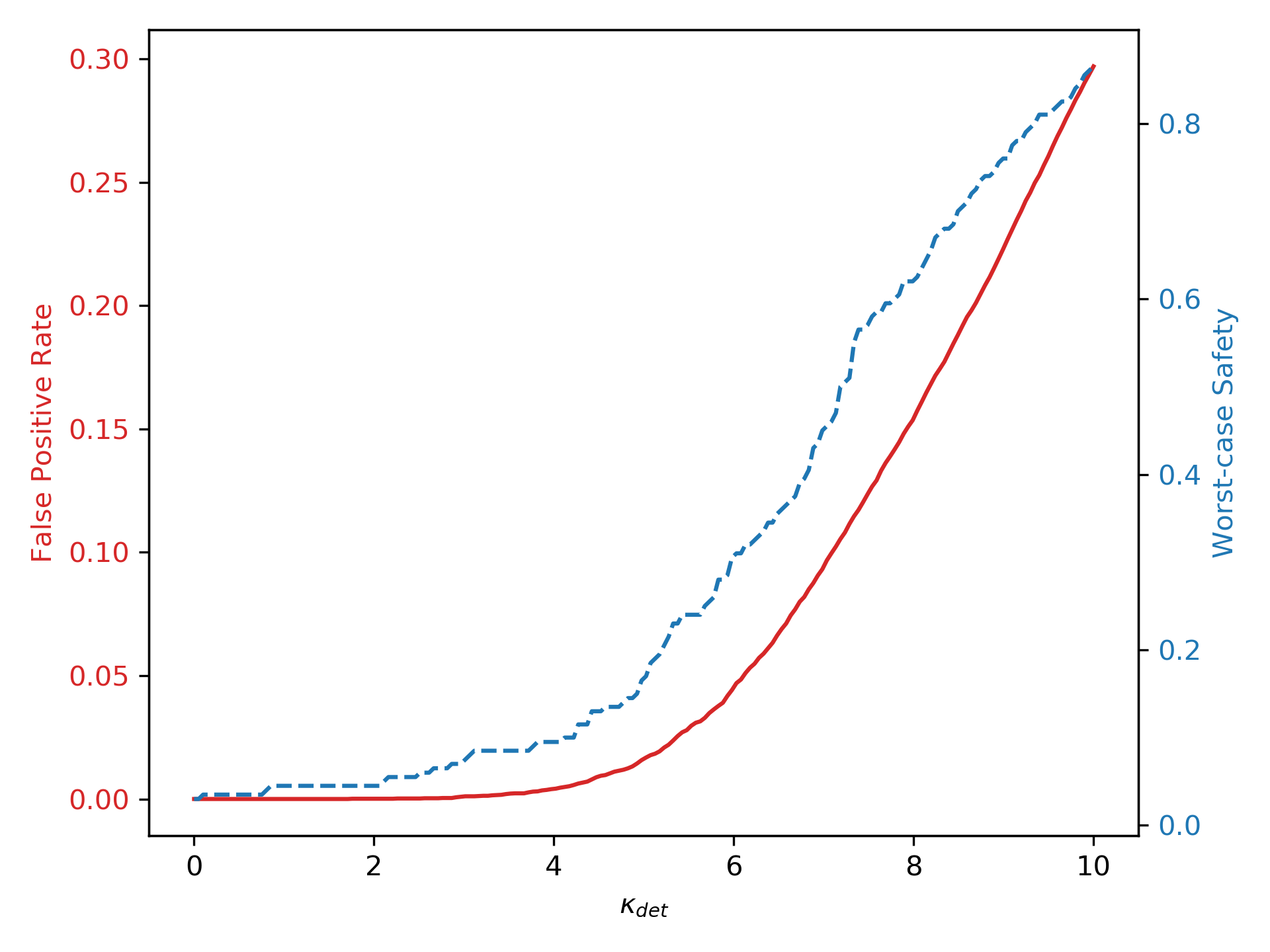}
			
			\caption{Variation of false positive rate and worst-case robustness with the $\kappa_{det}$ parameter. The blue axis represents the proportion of images out of a $200$ sample subset of the CIFAR-10 test set, for which a black-box, unrestricted PGD attacker cannot find a single location of the patch that evades detection.}
			\label{fig:worst_case_tradeoff}
			
		\end{center}
	\end{figure}

	We consider an adversary that suspects that there is a detection method in place, but has no information (and does not wish to make any assumptions) about -- nor can they query -- the detector output. We place this adversary in the black-box category, even though they are borderline gray-box. A feasible attack strategy in this case is to brute-force the location of the patch in the image (e.g., spamming a face tagging system using clone accounts). We evaluate the worst-case performance of the detector against this adversary: if even a single location in an image leads to a missed detection, we consider the entire image compromised. The results are shown in \figurename{ \ref{fig:worst_case_tradeoff}}, where it can be seen that using the default value of $\kappa_{det} = 3$ leads to a worst-case detection rate of $8\%$.Increasing $\kappa_{det}$ to $8$ increases the false positive rate to $17\%$, but ensures that, on average, half of the test samples are protected against all possible patch locations attempted by a black-box adversary.
	
	\subsubsection{Single-Pixel Attack}
	
	We test the performance of our approach against the single-pixel attack \cite{su2019one}. We run a targeted attack on $300$ correctly classified images in the test, for each other possible class targeted, for a total of $2700$ attacks, each for $200$ iterations (generations) and a population size of $75$. A number of $300$ attacks are successful in finding an adversary and only two evade detection with $\kappa_{det} = 3$. Interestingly, even without using a confidence threshold, $60.33\%$ of the successful single-pixel attacks are detected by the detector $H$ itself.

	\subsection{Adaptive (White-Box) Attacks}
	We evaluate the robustness of the residual detection method against a C\&W adversary that has complete knowledge of the model, including the parameter $\sigma$ used for performing wavelet denoising, the confidence threshold $\kappa_{det}$, and the detector weights $\theta_H$.	Since the wavelet denoising block is non-differentiable, we apply the straight-through approach \cite{athalye2018obfuscated} to estimate the hidden gradients, by exactly computing the residual during the forward pass, and approximating its gradient with unit value during the backward pass. The wavelet denoising operation takes up a large part of the complexity of this attack. For this reason, we perform it only every five iterations, since we find that this does not hinder optimization.
	
	We run our attacks on a detector with $\kappa_{det} = 3$ and attack a set of $256$ correctly classified test images, with a targeted attack to another random label. We run $2000$ iterations per binary search step for the restricted attack, for a number of $7$ steps, and we run $10000$ iterations for the unrestricted attack.	Table \ref{table:wbox} presents the results in terms of success rate, average and worst-case $L_2$ distances, for the $L_2$-constrained and unrestricted white-box attacks. We note that the average success rate is decreased when comparing to a black-box adversary and the worst-case distortion increases by a factor of $1.5$ times as well. Finally, the worst-case performance counts a sample as compromised if at least one of the $40$ patch locations bypasses detection. The success rate of this attacker is close to $100\%$, but with an increased distortion cost of $0.344$. One caveat here is that we did not search across all possible patch locations, but only $40$ out of $676$, meaning it is likely possible to increase this success rate to exactly $100\%$ and the true worst-case value of the required distortion.
	
	\begin{table}[t]
		\caption{Success Rate (SR) and perturbation distance for white-box attacks on the proposed detection method with $\kappa_{det} = 3$.}
		\label{table:wbox}
		\begin{center}
			
					\begin{tabular}{|l|c|c|c|r|}
						\hline
						Attack & C\&W $L_2$ & C\&W, Unrestricted \\
						\hline
						Avg. SR       & $22.82$\% & $32.05$\% \\
						Worst-Case SR    & $72.65$\% & $87.8$\% \\
						Average $L_2$      & $0.3$ & $0.375$ \\
						Worst-Case $L_2$   & $0.286$ & $0.344$ \\
						\hline
					\end{tabular}

		\end{center}
		\vskip -0.1in
	\end{table}
	
	\subsection{Gray-Box Attacks}
	Finally, we investigate the transferability by assuming an adversary has complete knowledge about the datasets used to train and validate the performance of the detection scheme, as well as the hard labels output by the detector during training, but not testing phase. We train a deep convolutional network with four convolutional layers to mimic the detector as a substitute model, trained on the training set and the predicted labels of the detector. We use the same $L_2$ weight regularization factor of $5\times 10^{-3}$. Then, we generate high confidence white-box adversarial examples for the substitute classifier-detector ensemble. Training the substitute model is successful, with a validation accuracy of $98\%$ on the same data used by the detector. When testing, we generate $40$ square patches for $256$ test images and obtain an attack transfer rate of $100\%$ on the classifier itself, but only a $15.22\%$ rate on the classifier-detector ensemble. We thus conclude that our method resists the transfer of high-confidence examples.
	
	\section{Conclusions}
	We have investigated the problem of detecting adversarial samples generated by patch adversarial attacks in an attempt to more closely match threat models that may arise in practical situations. Our proposed solution uses the residual high-frequency content of an image to distinguish between clean and attacked samples. We have experimentally shown that our method generalizes to strong black-box adversaries, resists transfer attacks, and decreases the success rate of white-box adversaries.
	Upon visually inspecting the images output by an adaptive adversary, we make one interesting observation: even though the required distortion increases, the patches have smoother textures and color gradients. Thus, these represent almost \textit{natural} adversarial examples that bypass our wavelet-based scheme. Future research directions deal with combining our detector with other criteria to eliminate these blind spots.
	
	\bibliographystyle{IEEEtran}
	\bibliography{paper}
	
\end{document}